\documentclass[runningheads]{llncs}
\usepackage{cite}
\usepackage{booktabs}
\usepackage{multirow}
\usepackage{color}
\usepackage{graphicx}
\usepackage{url}
\begin{document}
\setlength{\tabcolsep}{4pt}
\title{Deep Learning versus Classical Regression for Brain Tumor Patient Survival Prediction}	
\titlerunning{DL vs. Classical Regression for Brain Tumor Patient Survival Prediction}
%
\author{Yannick  Suter\inst{1}\and
Alain Jungo\inst{1}\and
Michael Rebsamen\inst{1}\and
Urspeter Knecht\inst{1}\and
Evelyn Herrmann\inst{2}\and
Roland Wiest\inst{3}\and
Mauricio Reyes\inst{1}}
\authorrunning{Yannick Suter et al.}

\institute{Institute for Surgical Technology and Biomechanics, University of Bern, Bern, Switzerland\\
\email{yannick.suter@istb.unibe.ch}
\and
University Clinic for Radio-oncology, Inselspital, Bern University Hospital, University of Bern, Bern,
Switzerland
\and
Support Center for Advanced Neuroimaging, University Institute of Diagnostic and
Interventional Neuroradiology, Inselspital, University of Bern, Switzerland
}
\maketitle              
\begin{abstract}
Deep learning for regression tasks on medical imaging data has shown promising results. However, compared to other approaches, their power is strongly linked to the dataset size. In this study, we evaluate 3D-convolutional neural networks (CNNs) and classical regression methods with hand-crafted features for survival time regression of patients with high-grade brain tumors.
The tested CNNs for regression showed promising but unstable results. The best performing deep learning approach reached an accuracy of $51.5\%$ on held-out samples of the training set. All tested deep learning experiments were outperformed by a Support Vector Classifier (SVC) using 30 radiomic features. The investigated features included intensity, shape, location and deep features.

The submitted method to the BraTS 2018 survival prediction challenge is an ensemble of SVCs, which reached a cross-validated accuracy of $72.2\%$ on the BraTS 2018 training set, $57.1\%$ on the validation set, and $42.9\%$ on the testing set.

The results suggest that more training data is necessary for a stable performance of a CNN model for direct regression from magnetic resonance images, and that non-imaging clinical patient information is crucial along with imaging information.

\keywords{Brain tumor \and Survival prediction \and Regression \and 3D-Convolutional Neural Networks}
\end{abstract}
\section{Introduction}
High-grade gliomas are the most frequent primary brain tumors in humans. Due to their rapid growth and infiltrative nature, the prognosis for patients with gliomas ranking at grade III or IV on the Word Health Organization (WHO) grading scheme \cite{who2016} is poor, with a median survival time of only 14 months. Finding biomarkers based on magnetic resonance (MR) imaging data could lead to an improved disease progression monitoring and support clinicians in treatment decision-making \cite{gillies2015radiomics}.

Predicting the survival time from pre-treatment MR data is inherently difficult, due to the high impact of the extent of resection (e.g., \cite{sanai2011extent,meier2017automatic}) and response of the patient to chemo- and radiation therapy. The progress in the fields of automated brain tumor segmentation and radiomics have led to many different approaches to predict the survival time of high-grade glioma patients. Further, the introduction of the survival prediction task in the BraTS challenge 2017 (\cite{Menze2015TheBRATS}, \cite{2018arXiv181102629B}) makes a direct performance comparison of methods possible. The current state-of-the-art approaches can roughly be classified into 
\begin{enumerate}
	\item Classical radiomics: Extracting intensity features and/or shape properties from segmentations and use regression techniques such as random forest (RF) regression \cite{breiman1984classification}, logistic regression, or sparsity enforcing methods such as LASSO \cite{tibshirani1996regression}.
	\item Deep features: Neural networks are used to extract features, which are subsequently fed into a classical regression method such as logistic regression \cite{cox1958regression}, support vector regression (SVR), or support vector classification (SVC) \cite{lampert2009kernel}.
	\item A combination of classical radiomics and deep features (e.g., \cite{lao2017deep}).
	\item Survival regression from MR data using deep convolutional neural networks (CNNs) with or without additional non-imaging input (e.g., \cite{li2017deep}).
\end{enumerate}

Our experiments with 3D-CNNs for survival time regression confirmed observations made by other groups in last year's competition (e.g., \cite{li2017deep}), that these models tend to converge and overfit extremely fast on the training set, but show poor generalization when tested on the held-out samples. The top-ranked methods of last year's competition were mainly based on RF. A reason for this may be the relatively few samples to learn from. Classical regression techniques typically have fewer learnable parameters compared to a CNN and perform better with sparse training data.

We present experiments ranging from simple linear models to end-to-end 3D-CNNs and combinations of classical radiomics with deep learning to benchmark new, more sophisticated approaches against established techniques. We believe that a thorough comparison and discussion will provide a good baseline for future investigations of survival prediction tasks.

\section{Methods}

\subsection{Data}
The provided BraTS 2018 training and validation datasets for the survival prediction task consist of 163 and 53 subjects, respectively. The challenge ranking is based on the performance on a test dataset with 77 subjects with gross total resection (GTR). 

A subject contains imaging and clinical data. The imaging data includes images from the four standard brain tumor MR sequences (T1-weighted (T1), T1-weighted post-contrast (T1c), T2-weighted, and T2-weighted fluid-attenuated inversion-recovery (FLAIR)). All images in the datasets are resampled to isotropic voxel size (1$\times$1$\times$1 mm\textsuperscript{3}), size-adapted to 240$\times$240$\times$155 mm\textsuperscript{3}, skull-stripped, and co-registered. The clinical data comprises the subject's age and resection status. The three possible resection statuses are: (a) gross total resection (GTR), (b) subtotal resection (STR), and (c) not available (NA).

\subsubsection{Segmentation:} 
\label{sec:segmentation}
For our experiments, we rely on segmentations of the three brain tumor sub-compartments (i.e., enhancing tumor, edema, and necrosis combined with non-enhancing tumor). In the validation and testing dataset, the segmentation is not provided due to the overlap with the data of the BraTS 2018 segmentation task. To obtain the required segmentations, we thus employ the cascaded anisotropic CNN by Wang et al. \cite{wang2017automatic}. Their method is publicly available\footnote{\url{https://github.com/taigw/brats17}} and contains pre-trained models on the BraTS 2017 training dataset, which is identical to the BraTS 2018 \cite{Bakas2017AdvancingFeatures, Bakas2017SegCollectionGBM, Bakas2017SegnCollLGG} training dataset. This enables us to compute the segmentations with the available models without retraining a new segmentation network. 

\subsection{Deep Survival Prediction and Deep Features:}
\begin{figure}[ht]
\centering
\includegraphics[width=\textwidth]{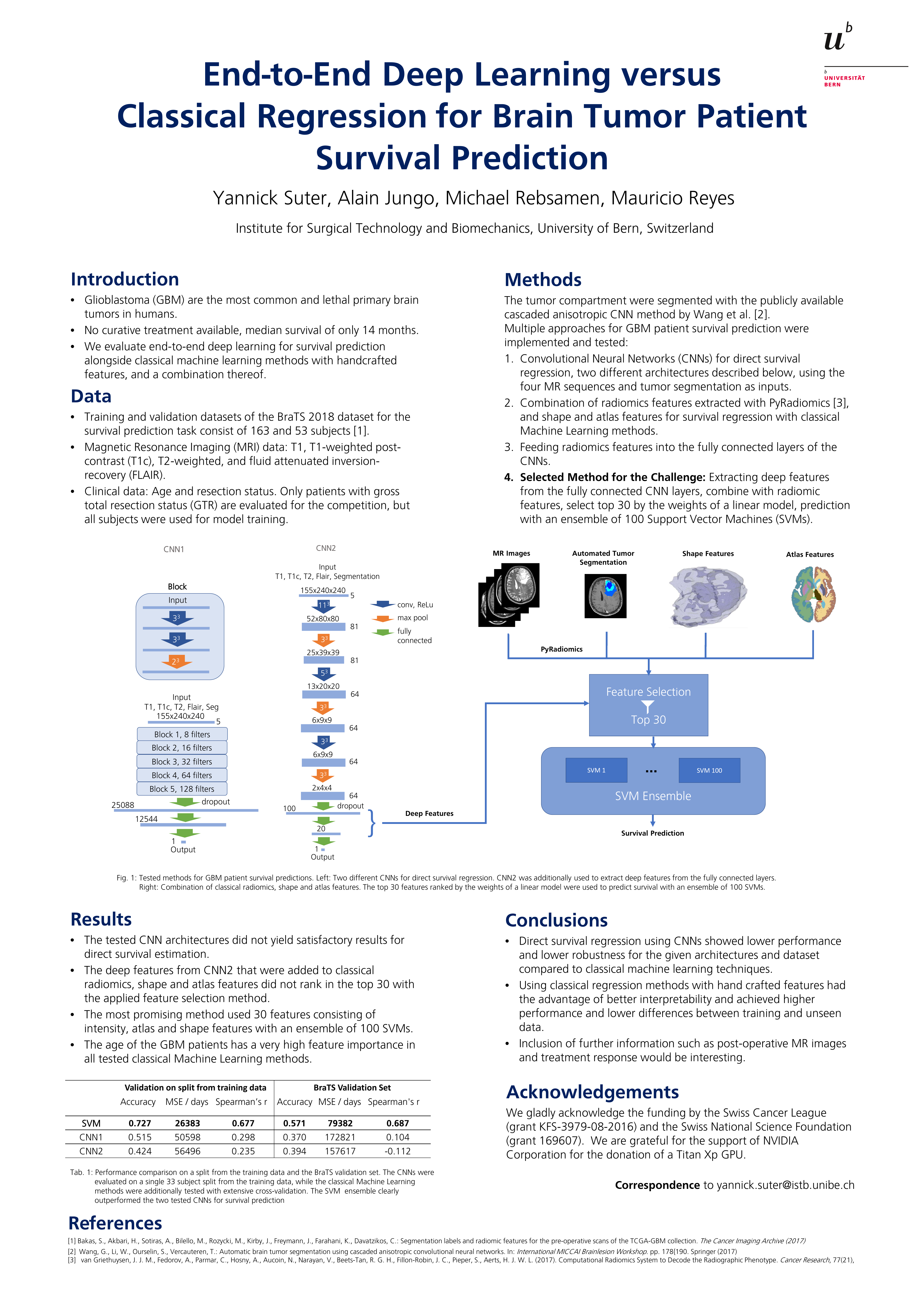}
\caption{Summary of the tested methods for GBM patient survival predictions. Left: The architectures of our CNNs for direct survival regression. CNN2 was additionally used to extract deep features from the fully connected layers. For direct regression, the two last fully connected layers of CNN2 had 2048 and 384 elements.
Right: Combination of classical radiomics, shape, and atlas features. The top 30 features were used to predict survival classes with a SVC.}
\label{fig:cnns}
\end{figure}

Two different CNNs are built for the survival regression task (see Figure \ref{fig:cnns}). 
CNN1 consists of five blocks with an increasing number of filters, each block has two convolutional layers and a max pooling operation. The last block is connected to two subsequent fully connected layers.
CNN2 consists of three convolutional layers with decreasing kernel sizes with intermediary max-pooling, followed by fully-connected layers connected to the single value regression target. To include clinical information into the CNN2, the age and resection status were appended to the first fully-connected layers of CNN2, which we refer to as CNN2+Age+RS.

Both CNN variants take the four MR sequences and additionally the corresponding segmentation (see \ref{sec:segmentation}) as input, and output the predicted survival in days. We observed no performance gain by the additional segmentation input but it improved the training behavior of the network. Instead of regressing the survival days, we also tested direct classification in long-, mid-, and short-term survival, but without improvements. 

We trained the CNNs with the Adam optimizer \cite{kinga2015method} and a learning rate of 10\textsuperscript{-5}, and performed model selection based on Spearman's rank coefficient on a held-out set. Batch normalization and more dropout layers did not lead to improvements, neither on the training behaviour nor the results.

\subsubsection{Deep feature extraction:} For the extraction of deep features, the size of the two last fully connected layers are decreased to 100 and 20 elements. The activations of these two layers serve as deep feature sets.

\subsection{Classical Survival Prediction}
\subsubsection{Feature extraction:} We extract an initial set of 1353 survival features from the computed segmentation together with the four MR images (i.e., T1, T1c, T2, and FLAIR).
\subsubsection{Gray-level and basic shape:}
1128 intensity and 45 shape features are computed with the open-source Python package \textit{pyradiomics}\footnote{\url{https://github.com/Radiomics/pyradiomics}} version 2.2.0 \cite{pyrad2017}. It includes shape, first-order, gray level co-occurrence matrix, gray level size zone matrix, gray level run length matrix, neighbouring gray tone difference matrix, and gray level dependence matrix features. Z-score normalization and a Laplacian of Gaussian filter with $\sigma=1$ is applied to the MR images before extraction. A bin width of 25 is selected and the minimum mask size set to 8 voxels. The features are calculated from all MR images and for all tumor sub-compartments provided by the segmentation (i.e., enhancing tumor, edema, necrosis combined with non-enhancing tumor).
\subsubsection{Shape:}
15 additional enhancing tumor shape features previously used as predictors for survival \cite{Perez-Beteta2017Glioblastoma:Study, jungo2017} complement the basic shape features from \textit{pyradiomics}. These features are the rim width of the enhancing tumor, geometric heterogeneity, combinations of rim width quartiles and volume ratios of all combinations of the three tumor compartments.
\subsubsection{Atlas location:} Tumor location has previously been used for survival prediction (e.g., \cite{Awad2017location}), therefore atlas location features are included. Affine registration is used to align all subjects to FreeSurfer's \cite{fischl2012freesurfer} \textit{fsaverage} subject and its subcortical segmentation (\textit{aseg}) is used as the atlas. The volume fraction of each anatomical region occupied by the contrast enhancing tumor is used as a feature, resulting in 43 features in total.
\subsubsection{Clinical information:}
The two provided clinical features resection status and age are further added to the feature set.

\subsubsection{Feature selection:}
Since the number of extracted features (n=1353) is much higher than the available samples (n=163), a subset of features needs to be used. Apart from being necessary for many machine learning methods, a reduction of the feature space improves the interpretability of possible markers regarding survival \cite{PEREIRA2018228}. 

We analyzed the following feature selection techniques to find the most informative features: (a) step wise forward/backward selection with a linear model, (b) univariate feature selection, and (c) model-based feature selection by the learned feature weights or importances. We observed a rather low overlap among the selected features by the different techniques, or even the parameterization of the techniques. Consequently, we chose the feature subsets according to their performance on the training dataset for different classical machine learning methods (e.g., linear regression, SVC, and RF). The best results were obtained by the feature subset produced by the model-based feature selection from a sparse SVC model, which consists of the features listed in Table~\ref{tab:feat}.

Our model-based feature selection identified age by far as most important feature. Additionally, a majority of the 30 selected features are intensity-based, but the subset also contains shape and atlas features. We note that none of the 120 deep features was retained.

\begin{table}[ht!]\centering
\caption{Selected feature set with feature category (Cat.), tumor sub-compartment (Comp.), and the MR image (Img.) in decreasing order of importance. ED: Edema, ET: Enhancing tumor, NCR/NET: Necrosis and non-enhancing tumor. Feature importance is decreasing from top to bottom.}
\label{tab:feat}
\begin{tabular}{llll}
\toprule 
\textbf{Feature} & \textbf{Cat.} & \textbf{Comp.} & \textbf{Img.} \\
\midrule
Age & Clinical & & \\
Sphericity & Shape & ED & \\
Optic Chiasm & Atlas & \\
Small Area Low Gray Level Emphasis & Intensity & ED & T2 \\
Correlation & Intensity & CE & Flair \\
Cluster Shade & Intensity & NCR/NET & T1c \\
Small Dependence High Gray Level Emphasis & Intensity & CE & T1c \\
Correlation & Intensity & NCR/NET & T1 \\
Maximum & Intensity & ED & T1 \\
Maximum & Intensity & ED & T1c \\
Left Amygdala & Atlas & & \\
Information Measure of Correlation 1 & Intensity & NCR/NET & T1 \\
Large Dependence Low Gray Level Emphasis & Intensity & NCR/NET & T1c \\
Cluster Shade & Intensity & ED & T2 \\
Inverse Variance & Intensity & ED & T1 \\
Small Dependence High Gray Level Emphasis & Intensity & CE & T2 \\
Median & Intensity & CE & T2 \\
Busyness & Intensity & ED & T1 \\
Correlation & Intensity & NCR/NET & T1c \\
Right-vessel & Atlas & & \\
Large Area Low Gray Level Emphasis & Intensity & ET & Flair \\
Right Caudate & Atlas & & \\
Difference Variance & Intensity & ED & Flair \\
Right Cerebellum Cortex & Atlas & & \\
Cluster Prominence & Intensity & ED & T2 \\
Maximum 2D Diameter Slice & Intensity & ED & \\
Inverse Difference Normalized & Intensity & CE & Flair \\
Skewness & Intensity & ED & Flair \\
Median & Intensity & ET & T1 \\
Right Ventral Diencephalon & Atlas & & \\
\bottomrule
\end{tabular}
\end{table}

\subsubsection{Feature-based models:}
Although the BraTS survival prediction task is set up as a regression task, the final evaluation is performed on the classification accuracy of the three classes: short-term (less than 10 months), mid-term (between ten and 15 months), and long-term survivors (longer than 15 months). As a consequence, we include classification models in addition to the regression models in our experiments. Since the prediction is required in days of survival, the output of the classifiers needs to be transformed from a class (i.e., short-term, mid-term, long-term) to a day scalar. We do this by replacing each class by its mean time of survival (i.e. 147, 376, 626 days).

For our experiments, we consider the following feature-based regression and classification models \cite{friedman2001elements}:

\begin{minipage}{\linewidth}
\begin{itemize}
    \item Linear and logistic regression
    \item RF regression and classification 
    \item SVR and SVC
    \item SVC ensemble
\end{itemize}
\end{minipage}

We use 50 trees and an automatic tree depth for the RF models and linear kernels for the support vector approaches, SVR, and SVC. To handle the multi-class survival problem we employ the \textit{one-versus-rest} binary approach for SVC and logistic regression. The ensemble method consists of 100 SVC models that are separately built on random splits of 80\% of the training data. The final class prediction is performed by majority vote. We choose an ensemble to increase robustness against outliers or unrepresentative subjects in the training set. All classical feature-based models are implemented with \textit{scikit-learn}\footnote{\url{http://scikit-learn.org/stable/index.html}} version 0.19.1.

\subsection{Evaluation}
\label{sec:evaluation}
We evaluated the classical feature-based approaches by 50 repetitions of a stratified five-fold cross-validation on the BraTS 2018 training dataset. These repetitions allowed us to examine the models' robustness besides their average performance. The CNN approaches were evaluated on a randomly defined held-out split of the training set, consisting of 33 subjects. This held-out set was also used to evaluate a subset of the feature-based methods in order to compare classical approaches to the CNN approaches. Moreover, the classical and CNN models were evaluated on the BraTS 2018 validation set. This dataset contains 53 subjects but only the 28 subjects with resection status GTR are evaluated. Finally, we selected the best-performing model to predict survival on the BraTs 2018 challenge test dataset, which consists of 77 evaluated subjects with GTR resection status (out of 130 subjects).

\section{Results}
In this section, we compare the performance of the CNN to the classical feature-based machine learning models on the BraTS 2018 training and validation datasets, and present the BraTS 2018 test set results. We introduced a reference baseline for the comparison of the different models. This baseline consists of a logistic regression model solely trained on the age feature. This minimal model provides us with a reference for the training and validation set.

Table~\ref{tab:results_train} lists the results of the different models on the training dataset. To ensure a valid comparison, the table is subdivided by the two evaluation types, repeated cross-validation (CV) and hold-out (HO) (see Section \ref{sec:evaluation}). The results from the CV analysis highlights that by far the best results are achieved by the logistic regression, SVC, and ensemble SVC models, which performed very similarly. Except for the RF model, the classification models clearly outperformed their regression counterparts. The results from the HO analysis (Table~\ref{tab:results_train}, bottom) additionally reveals that well-performing classical methods (logistic regression and SVC) outperform all three CNN approaches (CNN1, CNN2, CNN2+Age+RS) by a large margin. 

\begin{table}[bh!]\centering
\caption{Results achieved on the BraTS 2018 training dataset by 100 stratified five-fold cross-validation (CV) runs (reported as mean$\pm$standard deviation) and on one split with 33 held-out (HO) samples. The baseline consists of a logistic regression model with age as single feature. Best results per metric and evaluation type (Eval.) are presented in bold. Acc.: Accuracy, MSE: Mean squared error, r\textsubscript{S}: Spearman's rank coefficient, RS: Resection status.}
\label{tab:results_train}
\begin{tabular}{clccc}
\toprule 
\textbf{Eval.} & \textbf{Method} &  \textbf{Acc.} & \textbf{MSE / days\textsuperscript{2}} & \textbf{r\textsubscript{S}}\\
\midrule
\multirow{8}{*}{CV} & Baseline & 0.489$\pm$0.06 & 136323$\pm$44378 & 0.300$\pm$0.14 \\
& Linear Regression & 0.552$\pm$0.08 & 260706$\pm$647176 & 0.573$\pm$0.12 \\
& SVR & 0.554$\pm$0.08 & 257542$\pm$637062 & 0.574$\pm$0.12  \\
& RF Regression & 0.444$\pm$0.08 & 117320$\pm$42503 & 0.332$\pm$0.17 \\
& Logistic Regression & 0.721$\pm$0.07 & \textbf{93158}$\pm$35665 & \textbf{0.617}$\pm$0.12 \\
& SVC & \textbf{0.722}$\pm$0.07 & 93571$\pm$35861 & 0.612$\pm$0.12 \\
& RF & 0.512$\pm$0.07 & 136334$\pm$47392 & 0.324$\pm$0.15  \\
& SVC Ensemble & 0.720$\pm$0.07 & 93485$\pm$35652 & \textbf{0.617}$\pm$0.12 \\
\midrule
\multirow{5}{*}{HO} & Logistic Regression & 0.697 & 30756 & 0.579 \\
& SVC & \textbf{0.727} & \textbf{28226} & \textbf{0.616} \\
& CNN1 & 0.515 & 50598 & 0.298 \\
& CNN2 & 0.424 & 56496 & 0.235 \\ 
& CNN2+Age+RS & 0.394 & 61798 & -0.194 \\
\bottomrule
\end{tabular}
\end{table}

Table~\ref{tab:results_valid} presents the results obtained on the validation dataset. We can observe similar patterns as for the training set results: the classification models outperform the regression models with respect to the accuracy (except the RF), the SVC models (i.e., SVC ensemble and SVC) achieve the best performances, and the CNNs remain behind the feature-based methods and the baseline. Additionally, we observe an overall decrease in performance compared to the training set results. 

\begin{table}[bh!]\centering
\caption{Results achieved on the BraTS 2018 validation dataset (28 samples). The baseline consists of a logistic regression model with age as single feature. Best results per metric are presented in bold. Acc.: Accuracy, MSE: Mean squared error, r\textsubscript{S}: Spearman's rank coefficient, RS: Resection status.}
\label{tab:results_valid}
\begin{tabular}{p{0.3\linewidth} cccc}
\toprule 
\textbf{Method} &  \textbf{Acc.} & \textbf{MSE / days\textsuperscript{2}} & \textbf{r\textsubscript{S}}\\
\midrule
Baseline & 0.464 & 128841 & 0.288 \\
Linear Regression & 0.464 & 89059 & 0.426 \\
SVR & 0.464 & 89035 & 0.426 \\
RF Regression & 0.393 & 80980 & 0.342 \\
Logistic Regression & 0.5 & 90791 & 0.393 \\
RF & 0.357 & 169782 & -0.058 \\
SVC & 0.536 & 85471 & 0.501 \\
SVC Ensemble  & \textbf{0.571} & \textbf{79381} & \textbf{0.556} \\
CNN1 & 0.370 & 172821 & 0.104\\
CNN2 & 0.394 & 157617 & -0.112\\ 
CNN2+Age+RS & 0.444 & 137912 & -0.005\\		
\bottomrule
\end{tabular}
\end{table}

The results of CNN1 on our validation split (accuracy of 0.515) could not be replicated on the BraTS validation set, where it performed poorly with an accuracy of 0.37. CNN2 showed worse results on our validation split than the deeper CNN1, but performed better on the BraTS validation set.

Overall, the SVC ensemble performed best on the training and validation set and we consequently selected it for the challenge, where our method achieved an accuracy of 0.429, a mean squared error of 327725 days\textsuperscript{2} and a Spearman's rank coefficient of 0.172.

\section{Discussion}
In this section, we discuss the presented results and highlight findings from the deep learning, and classical regression and classification experiments.

\subsubsection{CNNs:}
The two CNNs overfit very fast on the training data, and showed highly variable performance between epochs. Model selection during training was therefore challenging, since both the accuracy and Spearman's rank coefficient were very unstable. 

We postulate that more data would be needed to fully benefit from direct survival estimation with 3D-CNNs. When inspecting the filters of CNN1 and CNN2, most of the learning took place at the fully connected layers and almost none at the first convolutions layer. This effect and the fast overfitting of the CNN models indicate the lack of samples and are reasons for the poor performance on unseen data.

\subsubsection{Classical regression and classification:} Using classical regression techniques with hand-crafted features has the advantage of better interpretability. Models with fewer learnable parameters, such as the classical regression methods we tested, typically achieve more robust results on unseen data when only few training samples are available.

\begin{figure}[h!]
\centering
\includegraphics[width=\textwidth]{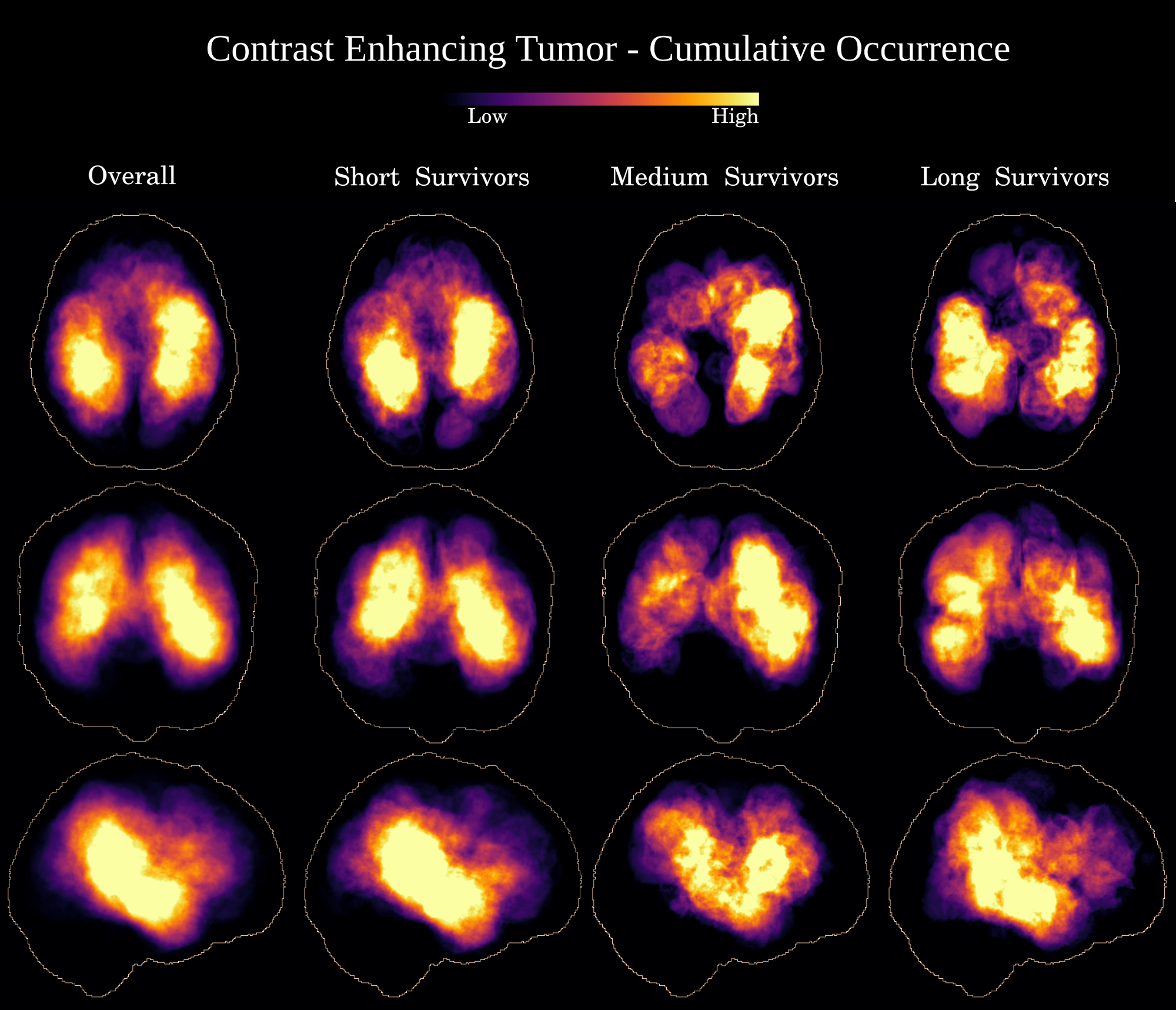}
\caption{Cumulative occurrence of the contrast enhancing tumor. Columns, from left to right: Overall across all three survival classes, short survivors ($<10$ months), medium survivors ($\geq 10$ months and $\leq 15$ months), and long survivors ($>15$ months). Rows: Projection along axial, coronal and sagittal axes.}
\label{fig:cetlocation}
\end{figure}

The atlas used for feature extraction most likely has too many regions for the number of training samples. Small anatomical structures, such as the optic chiasm, may not be accurately identified by simple registration to an atlas.
Figure \ref{fig:cetlocation} shows the distribution of the contrast enhancing tumor segmentation per survival class.
The short survivors with large contrast enhancing tumor loads contribute highly to the overall cumulative occurrence in the training data. The class-wise occurrence maps suggest that more training samples are needed to detect predictive location patterns (e.g. as reported in \cite{steed2016differential} and \cite{rathore2018radiomic}). Additionally, a coarser atlas subdivision driven by clinical knowledge is in order. In the light of this caveat, the location features used here should be seen as approximate localization information with limited clinical interpretability.

\subsubsection{Performance on the testing data:}
The accuracy of $72.2\%$ and $57.1\%$ on the training and validation set could not be maintained on the testing data. The large performance drop might be caused by still too many features compared to the training set size. Other possible reasons may include a lack of feature robustness or different class distribution compared to the training data.
Moreover, the survival time distributions within classes do not drop at the class boundaries, such that a small shift in the prediction can cause a large accuracy difference because ending up in a different class.

In conclusion, classical machine learning techniques using hand-crafted features still outperform deep learning approaches with the given data set size. The robustness of features regarding image quality and across MR imaging centers needs close attention, to ensure that the performance can be maintained on unseen data. We hypothesize that adding post-treatment imaging data and more clinical information to the challenge dataset would boost the performance of the survival regression.

\section{Acknowledgements}
We gladly acknowledge the support of the Swiss Cancer League (grant KFS-3979-08-2016) and the Swiss National Science Foundation (grant 169607). We are grateful for the support of the NVIDIA corporation for the donation of a Titan Xp GPU. Calculations were partly performed on UBELIX, the HPC cluster at the University of Bern.

\bibliographystyle{splncs04}
\bibliography{samplepaper}

\begin{thebibliography}{10}
\providecommand{\url}[1]{\texttt{#1}}
\providecommand{\urlprefix}{URL }
\providecommand{\doi}[1]{https://doi.org/#1}

\bibitem{Awad2017location}
Awad, A.W., Karsy, M., Sanai, N., Spetzler, R., Zhang, Y., Xu, Y., Mahan, M.A.:
  Impact of removed tumor volume and location on patient outcome in
  glioblastoma. Journal of Neuro-Oncology  \textbf{135}(1),  161--171 (Oct
  2017). \doi{10.1007/s11060-017-2562-1}

\bibitem{Bakas2017SegCollectionGBM}
Bakas, S., Akbari, H., Sotiras, A., Bilello, M., Rozycki, M., Kirby, J.,
  Freymann, J., Farahani, K., Davatzikos, C.: {Segmentation labels and radiomic
  features for the pre-operative scans of the TCGA-GBM collection}. The Cancer
  Imaging Archive  (2017). \doi{10.1038/sdata.2017.117}

\bibitem{Bakas2017SegnCollLGG}
Bakas, S., Akbari, H., Sotiras, A., Bilello, M., Rozycki, M., Kirby, J.,
  Freymann, J., Farahani, K., Davatzikos, C.: {Segmentation Labels and Radiomic
  Features for the Pre-operative Scans of the TCGA-LGG collection}. The Cancer
  Imaging Archive  (2017). \doi{10.1038/sdata.2017.117}

\bibitem{2018arXiv181102629B}
Bakas, S., Reyes, M., et~Int, Menze, B.: {Identifying the Best Machine Learning
  Algorithms for Brain Tumor Segmentation, Progression Assessment, and Overall
  Survival Prediction in the BRATS Challenge}. ArXiv e-prints  (Nov 2018)

\bibitem{Bakas2017AdvancingFeatures}
Bakas, S., Akbari, H., Sotiras, A., Bilello, M., Rozycki, M., Kirby, J.S.,
  Freymann, J.B., Farahani, K., Davatzikos, C.: {Advancing The Cancer Genome
  Atlas glioma MRI collections with expert segmentation labels and radiomic
  features}. Scientific Data  \textbf{4},  170117 (9 2017).
  \doi{10.1038/sdata.2017.117}

\bibitem{breiman1984classification}
Breiman, L., Friedman, J.H., Olshen, R.A., Stone, C.J.: Classification and
  regression trees  (1984)

\bibitem{cox1958regression}
Cox, D.R.: The regression analysis of binary sequences. Journal of the Royal
  Statistical Society. Series B (Methodological) pp. 215--242 (1958)

\bibitem{fischl2012freesurfer}
Fischl, B.: Freesurfer. Neuroimage  \textbf{62}(2),  774--781 (2012).
  \doi{10.1016/j.neuroimage.2012.01.021}

\bibitem{friedman2001elements}
Friedman, J., Hastie, T., Tibshirani, R.: The elements of statistical learning,
  vol.~1. Springer series in statistics New York, NY, USA: (2001).
  \doi{0.1007/b94608}

\bibitem{gillies2015radiomics}
Gillies, R.J., Kinahan, P.E., Hricak, H.: Radiomics: images are more than
  pictures, they are data. Radiology  \textbf{278}(2),  563--577 (2015).
  \doi{10.1148/radiol.2015151169}

\bibitem{pyrad2017}
van Griethuysen, J.J., Fedorov, A., Parmar, C., Hosny, A., Aucoin, N., Narayan,
  V., Beets-Tan, R.G., Fillion-Robin, J.C., Pieper, S., Aerts, H.J.:
  Computational radiomics system to decode the radiographic phenotype. Cancer
  research  \textbf{77}(21),  e104--e107 (2017).
  \doi{10.1158/0008-5472.CAN-17-0339}

\bibitem{jungo2017}
Jungo, A., McKinley, R., Meier, R., Knecht, U., Vera, L., P{\'e}rez-Beteta, J.,
  Molina-Garc{\'\i}a, D., P{\'e}rez-Garc{\'\i}a, V.M., Wiest, R., Reyes, M.:
  Towards uncertainty-assisted brain tumor segmentation and survival
  prediction. In: International MICCAI Brainlesion Workshop. pp. 474--485.
  Springer (2017). \doi{10.1007/978-3-319-75238-9\_40}

\bibitem{kinga2015method}
Kinga, D., Adam, J.B.: A method for stochastic optimization. In: International
  Conference on Learning Representations (ICLR). vol.~5 (2015)

\bibitem{lampert2009kernel}
Lampert, C.H., et~al.: Kernel methods in computer vision. Foundations and
  Trends{\textregistered} in Computer Graphics and Vision  \textbf{4}(3),
  193--285 (2009). \doi{10.1561/0600000027}

\bibitem{lao2017deep}
Lao, J., Chen, Y., Li, Z.C., Li, Q., Zhang, J., Liu, J., Zhai, G.: A deep
  learning-based radiomics model for prediction of survival in glioblastoma
  multiforme. Scientific reports  \textbf{7}(1),  10353 (2017).
  \doi{10.1038/s41598-017-10649-8}

\bibitem{li2017deep}
Li, Y., Shen, L.: Deep learning based multimodal brain tumor diagnosis. In:
  International MICCAI Brainlesion Workshop. pp. 149--158. Springer (2017).
  \doi{10.1007/978-3-319-75238-9\_13}

\bibitem{who2016}
Louis, D.N., Perry, A., Reifenberger, G., Von~Deimling, A., Figarella-Branger,
  D., Cavenee, W.K., Ohgaki, H., Wiestler, O.D., Kleihues, P., Ellison, D.W.:
  The 2016 world health organization classification of tumors of the central
  nervous system: a summary. Acta neuropathologica  \textbf{131}(6),  803--820
  (2016). \doi{10.1007/s00401-016-1545-1}

\bibitem{meier2017automatic}
Meier, R., Porz, N., Knecht, U., Loosli, T., Schucht, P., Beck, J., Slotboom,
  J., Wiest, R., Reyes, M.: Automatic estimation of extent of resection and
  residual tumor volume of patients with glioblastoma. Journal of neurosurgery
  \textbf{127}(4),  798--806 (2017). \doi{10.3171/2016.9.JNS16146}

\bibitem{Menze2015TheBRATS}
Menze, B.H., Jakab, A., Bauer, S., et~Int, Reyes, M., Van~Leemput, K.: {The
  Multimodal Brain Tumor Image Segmentation Benchmark (BRATS)}. IEEE
  Transactions on Medical Imaging  \textbf{34}(10),  1993--2024 (10 2015).
  \doi{10.1109/TMI.2014.2377694}

\bibitem{PEREIRA2018228}
Pereira, S., Meier, R., McKinley, R., Wiest, R., Alves, V., Silva, C.A., Reyes,
  M.: Enhancing interpretability of automatically extracted machine learning
  features: application to a rbm-random forest system on brain lesion
  segmentation. Medical Image Analysis  \textbf{44},  228 -- 244 (2018).
  \doi{10.1016/j.media.2017.12.009}

\bibitem{Perez-Beteta2017Glioblastoma:Study}
P{\'{e}}rez-Beteta, J., Mart{\'{i}}nez-Gonz{\'{a}}lez, A., Molina, D.,
  Amo-Salas, M., Luque, B., Arregui, E., Calvo, M., Borr{\'{a}}s, J.M.,
  L{\'{o}}pez, C., Claramonte, M., Barcia, J.A., Iglesias, L., Avecillas, J.,
  Albillo, D., Navarro, M., Villanueva, J.M., Paniagua, J.C., Martino, J.,
  Vel{\'{a}}squez, C., Asenjo, B., Benavides, M., Herruzo, I., Delgado, M.d.C.,
  del Valle, A., Falkov, A., Schucht, P., Arana, E., P{\'{e}}rez-Romasanta, L.,
  P{\'{e}}rez-Garc{\'{i}}a, V.M.: {Glioblastoma: does the pre-treatment
  geometry matter? A postcontrast T1 MRI-based study}. European Radiology
  (2017). \doi{10.1007/s00330-016-4453-9}

\bibitem{rathore2018radiomic}
Rathore, S., Akbari, H., Rozycki, M., Abdullah, K.G., Nasrallah, M.P., Binder,
  Z.A., Davuluri, R.V., Lustig, R.A., Dahmane, N., Bilello, M., et~al.:
  Radiomic mri signature reveals three distinct subtypes of glioblastoma with
  different clinical and molecular characteristics, offering prognostic value
  beyond idh1. Scientific reports  \textbf{8}(1), ~5087 (2018).
  \doi{10.1038/s41598-018-22739-2}

\bibitem{sanai2011extent}
Sanai, N., Polley, M.Y., McDermott, M.W., Parsa, A.T., Berger, M.S.: An extent
  of resection threshold for newly diagnosed glioblastomas. Journal of
  neurosurgery  \textbf{115}(1), ~3--8 (2011). \doi{10.3171/2011.2.JNS10998}

\bibitem{steed2016differential}
Steed, T.C., Treiber, J.M., Patel, K., Ramakrishnan, V., Merk, A., Smith, A.R.,
  Carter, B.S., Dale, A.M., Chow, L.M., Chen, C.C.: Differential localization
  of glioblastoma subtype: implications on glioblastoma pathogenesis.
  Oncotarget  \textbf{7}(18),  24899 (2016). \doi{10.18632/oncotarget.8551}

\bibitem{tibshirani1996regression}
Tibshirani, R.: Regression shrinkage and selection via the lasso. Journal of
  the Royal Statistical Society. Series B (Methodological) pp. 267--288 (1996)

\bibitem{wang2017automatic}
Wang, G., Li, W., Ourselin, S., Vercauteren, T.: Automatic brain tumor
  segmentation using cascaded anisotropic convolutional neural networks. In:
  International MICCAI Brainlesion Workshop. pp. 178--190. Springer (2017).
  \doi{10.1007/978-3-319-75238-9\_16}

\end{thebibliography}

\end{document}